\documentclass[10pt,twocolumn,letterpaper]{article}

\usepackage{array}
\usepackage{subfigure}
\usepackage{epsfig}
\usepackage{epstopdf}
\usepackage{booktabs}
\usepackage{algorithm}
\usepackage{algpseudocode}
\usepackage{wacv}
\usepackage{times}
\usepackage{xcolor}
\usepackage{epsfig}
\usepackage{graphicx}
\usepackage{amsmath}
\usepackage{amssymb}

\def\wacvPaperID{545} 

\wacvfinalcopy 

\ifwacvfinal
\fi

\ifwacvfinal
\usepackage[breaklinks=true,bookmarks=false]{hyperref}
\else
\usepackage[pagebackref=true,breaklinks=true,colorlinks,bookmarks=false]{hyperref}
\fi

\begin{document}

\title{Utilizing Every Image Object for Semi-supervised Phrase Grounding}

\author{Haidong Zhu \quad \quad Arka Sadhu \quad \quad Zhaoheng Zheng \quad \quad Ram Nevatia\\
University of Southern California\\
{\tt\small {\{haidongz,asadhu,zhaoheng.zheng,nevatia\}@usc.edu}} 
}

\maketitle

\begin{abstract} 
    Phrase grounding models localize an object in the image given a referring expression. 
    The annotated language queries available during training are limited, which also limits the variations of language combinations that a model can see during training.
    In this paper, we study the case applying objects without labeled queries for training the semi-supervised phrase grounding. 
    We propose to use learned location and subject embedding predictors (LSEP) to generate the corresponding language embeddings for objects lacking annotated queries in the training set. 
    With the assistance of the detector, we also apply LSEP to train a grounding model on images without any annotation.
    We evaluate our method based on MAttNet on three public datasets: RefCOCO, RefCOCO+, and RefCOCOg. 
    We show that our predictors allow the grounding system to learn from the objects without labeled queries and improve accuracy by 34.9\% relatively with the detection results. 
\end{abstract}

\section{Introduction}\label{sec:intro}
The task of phrase grounding is to localize entities referred to by the given natural language phrases.
This requires associating words and phrases from language modality to objects and relations in the visual modality.
Grounding plays an important role in the applications for language and vision, such as visual question answering \cite{antol2015vqa,das2018embodied,gordon2018iqa,li2018attentive,zhang2019interpretable}, image retrieval \cite{wang2018joint} and captioning \cite{anderson2018bottom,vinyals2015show}.

\begin{figure}
    \centering
    \begin{tabular}{cc}
        \includegraphics[width=0.45\linewidth]{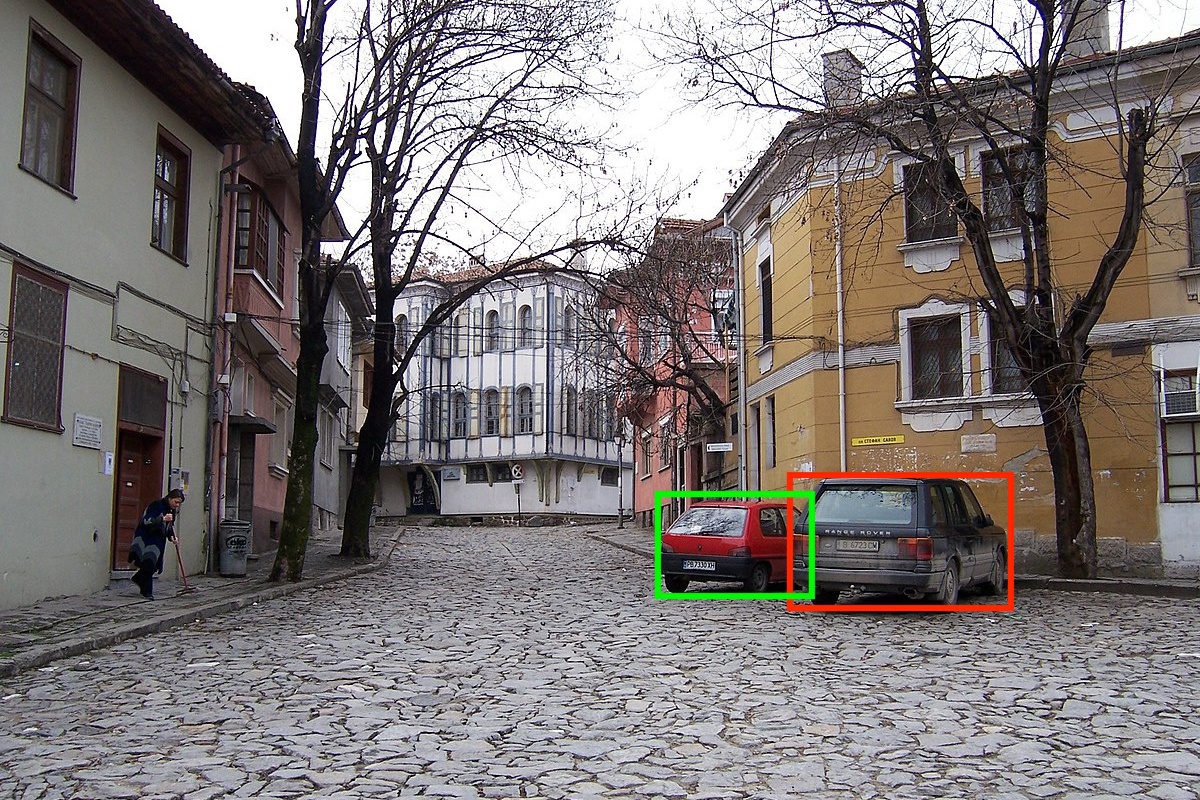} & 
        \includegraphics[width=0.45\linewidth]{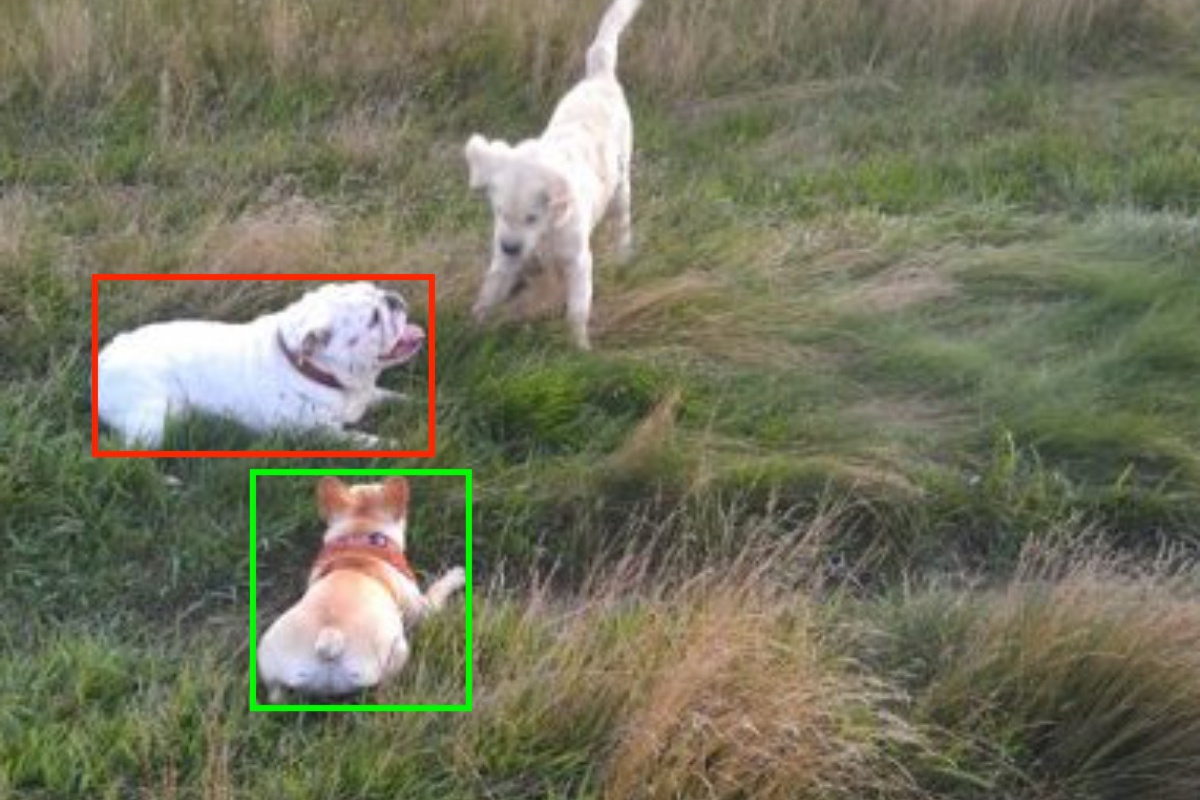} \\
         (a) &
        (b) \\
        
    \end{tabular}%
    
    \par
    \caption{Images in the training set where only some objects (those shown in red boxes) are labeled. Red boxes in the two images are annotated as `\textit{black car}' and `\textit{white dog lying on the grass on the left}' respectively, while objects in green boxes only have bounding boxes and category names, `\textit{car}' and `\textit{dog}', associated with them.}
    \label{fig:grounding example}
\end{figure}

We extend grounder training to a semi-supervised setting, where we assume objects are only sparsely annotated with language. 
Typical phrase grounding datasets only use a subset of the objects in the image with densely query annotations for training. Modern grounding systems, such as MAttNet \cite{yu2018mattnet}, are trained under fully supervised settings where every object used for training has a corresponding query. This limits the available data for training, where only images with dense annotations can be used. Fig.~\ref{fig:grounding example} shows two examples where some unlabeled objects cannot be used for training. Furthermore, the available queries limit the language variations that a model can see in the training set.

We propose a language embedding prediction module for unlabeled objects using knowledge from other objects in the training set, allowing us to use every image object for training. 
Previous semi-supervised grounding systems \cite{rohrbach2016grounding,liu2019knowledge}  still require full query annotations for the objects. 
To use the objects without labeled queries for training, simply using category names as the language queries is not effective;  the queries in a grounding dataset are discriminative within a category, whereas the category names alone do not perform such discrimination. 
Our method can directly be trained on images without queries annotations by generating the corresponding language embeddings. 


The embedding prediction module predicts query embeddings from visual information when the objects do not have associated phrases. 
In particular, we propose two embedding predictors to encode the subject and location properties from the given images. 
The predicted features are then combined with visual features to compute a grounding score. 
The predicted embeddings may not be perfect, but can still be useful in training. 
The predictors themselves are trained in an end-to-end framework; we call the resulting modules to be  Location and Subject Embedding Predictors, abbreviated as LSEP. 
We emphasize that LSEP modules are used only in the grounder training phase; at inference time, we always have the needed language query.

To  investigate the proposed semi-supervised setting, we use the following four-way characterization:
(i) sparsely annotated across images,
(ii) densely annotated fewer images,
(iii) only objects belong to a subset of the categories annotated, and
(iv) only certain supercategories, the parent categories, of objects are annotated. 
To create these settings we subsample three commonly used datasets: RefCOCO \cite{yu2016modeling}, RefCOCO+  and RefCOCOg \cite{mao2016generation}. 
Using MAttNet as our base architecture, we observe consistent improvement by adding LSEP modules using the labeled bounding boxes, as well as on images without any annotation with a detector.
 
In summary, our contributions are three-fold: 
(i) we introduce a new semi-supervised setting for phrase grounding training with limited labeled queries, 
(ii) we propose subject and location embedding predictors for generating related language features from the visual information to narrow the gap between supervised and semi-supervised tasks, and 
(iii) we extend the training of a semi-supervised phrase grounding model to unlabeled images with a detector.

\section{Related Work}\label{sec:rel}

\textbf{Phrase Grounding} has mainly been studied under the supervised setting, where query annotations are paired with the bounding boxes in the image.
On popular phrase grounding datasets like RefCOCO \cite{yu2016modeling, mao2016generation} and Flickr30k Entities \cite{flickrentitiesijcv, flickr30k}, state-of-the-art methods \cite{yu2018mattnet,Chen2017QueryGuidedRN,hu2017modeling,yu2018mattnet,li2019visualbert,su2019vl} use a two-step process: find all the objects using an object detector, such as Fast RCNN \cite{girshick2015fast} or Mask RCNN \cite{he2017mask}, and then jointly reason over the query and the detected objects.
QRC-Net \cite{Chen2017QueryGuidedRN} generates rewards for different proposals to match between visual and language modalities.
MAttNet \cite{yu2018mattnet} introduces an automatic weight evaluation method for different components of the query to match with proposals. 
Some research \cite{chen2019uniter,li2019visualbert,su2019vl,tan2019lxmert} try to apply attention \cite{liu2019improving} and visual-language transformer for cross-modality encoding between language queries and visual appearance. 
LXMert \cite{tan2019lxmert} applies the self-attention and cross-modality attention between the visual proposals and queries features to find the corresponding.
UNITER \cite{chen2019uniter} uses the pre-training system and greatly improve the performance for different visual-language tasks. 
Recently, single-stage grounding methods \cite{Yang2019AFA,sadhu2019zero}, where both steps are combined, have shown better performance on RefClef \cite{kazemzadeh2014referitgame}.
These models are pretrained on a large paired image-caption data like conceptual captions \cite{Sharma2018ConceptualCA} or aggregated vision+language datasets \cite{tan2019lxmert}.
As a result, it is difficult to evaluate them in the semi-supervised setting whose language annotations are scarce.
In our work, we build on MattNet \cite{yu2018mattnet} to enable semi-supervised learning.

\textbf{Semi-Supervised Phrase Grounding} focuses on the grounding problem where language annotations are scarce, which has not been explored extensively. 
However, there are some closely related publications on this subject. 
Some researchers \cite{rohrbach2016grounding,sadhu2019zero,chen2018knowledge,liu2019knowledge} apply visual language consistency for finding the entities in an image when the bounding boxes for the proposals are not available. 
GroundeR \cite{rohrbach2016grounding} considers semi-supervised localization where both language annotations and bounding boxes are present, while the association is provided only for a subset.
Zero-Shot Grounding (ZSG) \cite{sadhu2019zero}, in contrast, explores the grounding of novel objects for which bounding boxes are never seen.
Some other methods, such as \cite{xiao2017weakly}, use attention maps for finding the corresponding information without bounding boxes.
These semi-supervised methods still require language annotations for proposals during training.
Our proposed formulation assumes object bounding boxes are known, which can be easily acquired through the detection model, but language annotations for the detection boxes are not given. 
Compared with other existing semi-supervised grounding tasks, we can easily get the bounding boxes from a pretrained detector, while the missing of queries cannot be acquired without extra annotations.

\section{Method}\label{sec:met}
In this section, we introduce the phrase grounding task and MAttNet \cite{yu2018mattnet} in Sec.~\ref{bkg}, followed by the detailed descriptions of LSEP and discussions in Sec.~\ref{method} and Sec.~\ref{diss}.

\begin{figure*}
    \centering
    \begin{tabular}{cc}
        \includegraphics[width=0.45\linewidth]{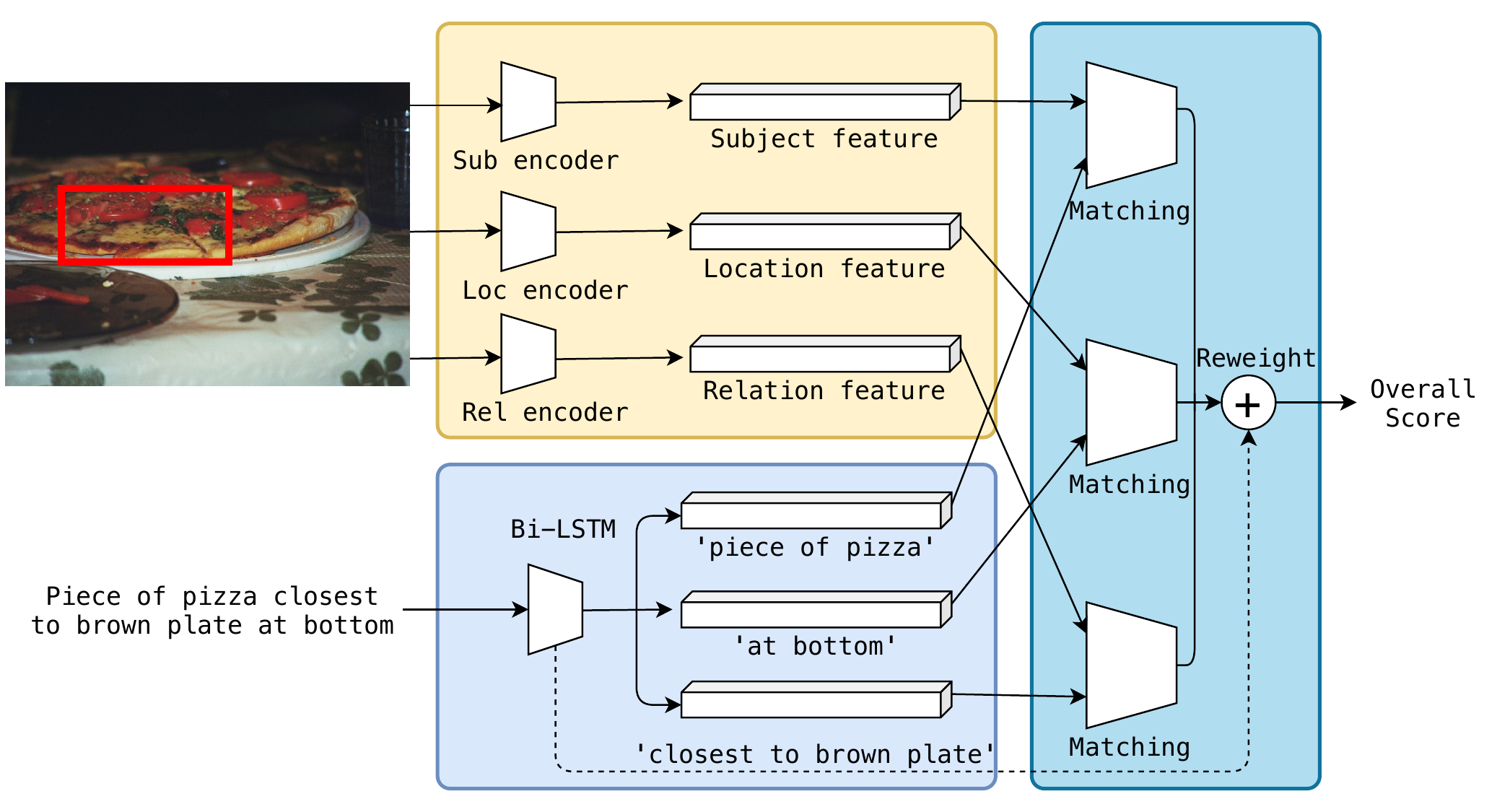} & 
        \includegraphics[width=0.45\linewidth]{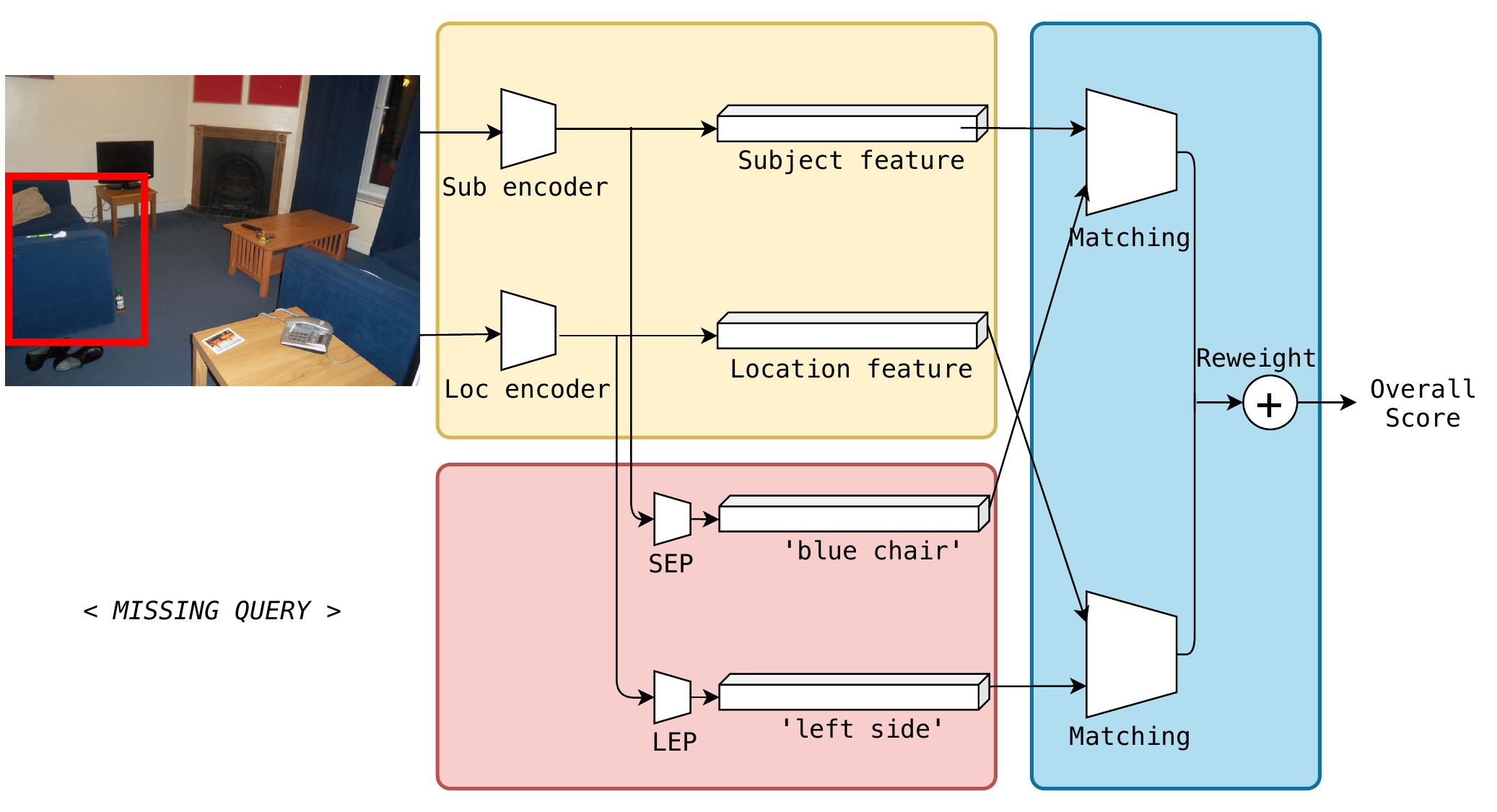} \\
       
        (a)& (b)\\
    \end{tabular}
     \caption{(a) MAttNet architecture  \cite{yu2018mattnet} used to generate query embeddings when descriptive phrases are available (b) Subject and Location Embedding predictors (SEP and LEP) which are used when only object category annotation is available. }
\label{fig:network}
\end{figure*}

\subsection{Background}\label{bkg}

A phrase grounding model takes a language query $q$ and the objects in the image $I$ as input. 
We use the enclosed bounding boxes to represent the objects. 
The bounding boxes are generated by a detector or come from groundtruth annotations, and the given query $q$ is matched to a specific bounding box in the image. 
The grounding model calculates the score of every bounding box $\{o_i\}_{i=1,...,n}$ and picks the box $o_q$ that best fits the given query $q$.  
In the supervised regime \cite{yu2018mattnet,rohrbach2016grounding,chen2018knowledge,chen2019uniter}, every bounding box used for training is annotated with a specific query. 

MAttNet splits a given query $q$ into three separate parts, subject $q^{subj}$, location $q^{loc}$ and relationship $q^{rel}$, by summing the one-hot vectors of corresponding words. MAttNet further predicts the relative weights, $w_{subj}$, $w_{loc}$ and $w_{rel}$, for every part in the query, and extracts subject features $v_{subj,i}$, location features $v_{loc,i}$ and relationship features $v_{rel,i}$ from the bounding boxes $\{o_i\}_{i=1,...,n}$ of the image $I$. 
The subject feature $v_{subj,i}$ contains the visual appearance and category information of $o_i$.
The location feature $v_{loc,i}$ contains the absolute position of bounding box $o_i$ and the relative position from its $N$ nearest neighbors. 
The relationship feature $v_{rel,i}$ includes both the relative position and visual representation of $N$ nearest proposals around $o_i$.
Then it calculates the similarity $S(o_i|q^{subj})$, $S(o_i|q^{loc})$ and $S(o_i|q^{rel})$ for every object-query pair.
MAttNet calculates the final score following 
\begin{eqnarray} \label{mattnet1}
S(o_i|r) =  w_{subj}S(o_i|q^{subj})&+w_{loc}S(o_i|q^{loc})\nonumber   \\
&+w_{rel}S(o_i|q^{rel})
\end{eqnarray}
During training, the model selects one positive pair $(o_i,q_i)$ along with two negative pairs, $(o_i,q_j)$ for negative query and $(o_k,q_i)$ for negative proposal, following \cite{yu2018mattnet} to optimize $L = L_{rank} + L_{subj}^{attr}$, where $L_{rank}$ is
\begin{eqnarray} \label{mattnet2}
L_{rank} =&\sum_i [\lambda_1 max(0, \Delta + S(o_i,q_j) - S(o_i,q_i)) \nonumber 
\\&+ \lambda_2 max(0, \Delta + S(o_k,q_i) - S(o_i,q_i))]
\end{eqnarray}
and $L_{subj}^{attr}$ is the cross-entropy loss for attribute prediction for $o_i$. 
We follow \cite{liu2017referring} to select the attributes. 
When a bounding box is not labeled with a query, MAttNet treats it as a negative visual pair $o_k$. During  inference, the optimal bounding box $\hat{o_i}$ is given by finding the maximum score in Eq.~\ref{mattnet1} 
among all the proposals to pair with the query $q$.

\subsection{Model Framework}\label{method}

Using MAttNet \cite{yu2018mattnet} as our backbone network, we propose two embeddings prediction modules: a subject embedding predictor and a location embedding predictor to generate language embeddings when corresponding queries are missing.
The complete LSEP framework is in Fig.~\ref{fig:network}.

\textbf{Subject Embedding Predictor}
The Subject embedding predictor consists of an encoder that maps visual input for subject $o_s$ to language dimension.
We follow  MAttNet \cite{yu2018mattnet} to extract the category feature $v_{cat}$ and attribute feature $v_{attr}$. 
Then we concatenate these two features for use as the visual embedding for subject $v_{subj} = (v_{attr}; v_{cat})$.
We use this module to transfer an existing attribute or embeddings for descriptive words to the known categories. 
We apply the following transformation
\begin{eqnarray*} \label{atttrans}
\Tilde{q}^{subj} = W_{subj}(v_{attr}; v_{cat})+ b_{subj}
\end{eqnarray*}
to generate the corresponding embedding $\Tilde{q}^{subj}$.
$W_{subj}$ and $b_{subj}$ are the weight and bias for the subject predictor. 
We use ";" to represent the concatenation between two different features.
During training, the grounding model takes $\Tilde{q}^{subj}$ as language input when ${q^{subj}}$ is missing.
$\Tilde{q}^{subj}$ lies in the same embedding space as ${q^{subj}}$ since we use ${q^{subj}}$ for supervision if it is available.
The subject embedding predictor transfers an embedding for attributes or descriptive words to the object without the full query to complete its attribute. 

\textbf{Location Embedding Predictor}
To generate the corresponding language embedding, we extract the absolute location of the bounding box as $loc_{a}$. 
Following \cite{yu2016modeling,yu2017joint}, we extract $N$ relative location features $loc_{r}$ for the $N$-nearest bounding boxes around it following \cite{yu2016modeling,yu2017joint}.
$loc_a$ is 5-D vector $loc_{a}$ in the form of  $[\frac{x_{min}}{w}, \frac{y_{min}}{h}, \frac{x_{max}}{w}, \frac{y_{max}}{h}, \frac{ Area_{B_i}}{Area_{I}}]$, 
and $loc_{r}$ is the concatenation of $N$ vectors in the form of  $[\frac{\Delta x_{min}}{w}, \frac{\Delta y_{min}}{h}, \frac{\Delta x_{max}}{w}, \frac{\Delta y_{max}}{h}, \frac{ Area_{B_i}}{Area_{B_j}}]$, representing the relative position between $N$ nearest objects and $o_i$. 
$w$ and $h$ represents the size of the image $I$; $x$ and $y$ represents the properties of the bounding boxes.
Location embedding predictor transfers the concatenation as $loc_{a}$ and $loc_{r}$ following
\begin{eqnarray*} \label{loctrans}
\Tilde{q}^{loc} = W_{loc}(loc_{a}; loc_{r} )+ b_{loc}
\end{eqnarray*}
to the language embedding $\Tilde{q^{loc}}$. 
$W_{loc}$ and $b_{loc}$ are the weight and bias for the location module.
By extracting visual embedding $v_{loc}$ following MAttNet \cite{yu2018mattnet}, the grounding model takes the $\Tilde{q}^{loc}$ as language input instead of ${q^{loc}}$ when it is not available in the semi-supervised setting. 

\subsection{Discussion}\label{diss}

In this subsection, we discuss some differences between LSEP and other existing methods, followed by the modified sampling policy, and why using the category names as queries doesn't help for semi-supervised phrase grounding.

\textbf{Differences with existing methods} 
Existing semi and weakly supervised grounding methods \cite{rohrbach2016grounding,xiao2017weakly,liu2019knowledge} focus on the circumstances where the bounding boxes are not available. 
They pair an image directly with a query and use attention maps to ground objects, thus cannot use objects without descriptive queries during training.
Since these methods cannot generate more variation of queries,
these methods are still limited to the word combinations. 
In LSEP, we do not require query annotations for every bounding box. 
The embedding predictors can generate the corresponding language embeddings from the visual embeddings.

\textbf{Sampling policy} For supervised grounding, all objects have descriptive queries associated with them. Thus, each object and query qualify as a positive or negative candidate. 
We apply a sampling strategy to accommodate the semi-supervised setting. 
In this setting, part of bounding boxes have descriptive phrases associated with them as the supervised setting, while others only have a category name associated. 
For positive samples, both type of samples qualify, but LSEP can use the predicted language embedding as positive examples. 
For negative queries, we sample two negative pairs $(o_i, q_j)$ and $(o_k, q_i)$ for every positive pair $(o_i, q_i)$. We sample the negative object $o_k$ and query $q_j$ from the available bounding boxes and queries, where the negative proposals and queries belong to the same object category as $o_i$ are preferred. 
When only the category name is available for $q_i$ or $q_j$, we avoid pairing it with the object from the same category as negative pairs, e.g., the car enclosed by the red box in Fig.~\ref{fig:grounding example} (a) cannot be used as a negative example for the query `car'.
In this case, proposals belong to different categories in the same image are preferred.

\textbf{Usage Analysis} 
When the bounding box only has the category name attached to it, an alternative to LSEP would be paring the bounding box with the query embedded generated by the category name. 
However, we find that using such an embedding is not helpful, and the use of LSEP gives significant improvements (as in Sec.~\ref{sec:res}). 
This improvement is due to additional discrimination provided by the two prediction modules. 
The descriptive queries are more effective at training a grounding system where the goal is to distinguish objects belonging to the same category. 

Consider an example where a bounding box containing a dog whose color is brown and it is to the left of the image, but the only annotation is its category name (`dog'). 
If an object is used as a positive example, the network must treat it the same as what the query describes, regardless of its location, color, or the similarities and differences of these entities compared with a negative query. 
Take the dog enclosed by the green bounding box in Fig.~\ref{fig:grounding example} (b) as an example. If we only use the category name `dog' as its positive query and use `white dog lying on the grass on the left' as the negative one, the grounding network cannot learn from the negative query that which part of the description is wrong, `white', `lying on the grass' or `on the left'. 
Predicting the subject and location embeddings provides a more accurate description for a more discriminative example. 
A similar analysis applies when we use the dog bounding box as a negative example; the network can better distinguish from the paired positive sample if the annotation is more specific. As each object box is selected as a positive sample once per epoch but not necessarily as a negative sample, the influence as a positive sample is much more critical.  

\section{Experiments}\label{sel:exp}
\subsection{Datasets}\label{dataset}
We use three public datasets for evaluations: RefCOCO \cite{yu2016modeling}, RefCOCO+  and RefCOCOg \cite{mao2016generation}, which are derived from MSCOCO \cite{lin2014microsoft} 2014 dataset. 
Queries in RefCOCO+ do not include absolute locations. 
MSCOCO \cite{lin2014microsoft} 2014 has 80 categories within 11 super categories.
A supercategory is the parent category of categories that share the same properties. For example, both `bus' and `car' belong to the supercategory `vehicle'.
We follow MAttNet \cite{yu2018mattnet} to create the training, validation and test sets. 
RefCOCO and RefCOCO+ include two test set split, testA and testB, where testA include the objects related with people, while testB include the objects that are irrelevant to people. RefCOCOg includes only one validation set and one test set.
We compare our method with other methods on two tasks: supervised and semi-supervised phrase grounding. 
For the semi-supervised task, we introduced four different data splits following i) annotation-based, ii) image-based, iii) category-based, and iv) supercategory-based strategies.

\textbf{Annotation-based selection} is to randomly choose objects from all the annotations in the training set. 
For objects remaining, only bounding boxes equipped with category names are available during training.  

\textbf{Image-based selection} is to select some images from the dataset and densely label the objects in these images with queries.  For the objects in the remaining images, we only have corresponding bounding boxes and category names for the objects. 
Labeled queries for the same image will be used or discarded together. 

\begin{table*}[]
\begin{center}
\resizebox{1.6\columnwidth}{!}
{
\begin{tabular}{p{1.45cm}p{2.8cm}<{\centering}p{0.6cm}<{\centering}p{0.6cm}<{\centering}p{0.6cm}<{\centering}p{0.0cm}<{\centering}p{0.6cm}<{\centering}p{0.6cm}<{\centering}p{0.6cm}<{\centering}p{0.0cm}<{\centering}p{0.6cm}<{\centering}p{0.6cm}<{\centering}}
\toprule
 Grounding&&\multicolumn{3}{c}{RefCOCO}
 && \multicolumn{3}{c}{RefCOCO+}
 && \multicolumn{2}{c}{RefCOCOg} \\ 
\cline{3-5}\cline{7-9}\cline{11-12}\\[-8pt]
Accuracy& type&Val & testA& testB &&Val& testA&testB &&Val& test  \\ \midrule   
MAttNet & annotation &  79.18 & 80.05 & 78.71  && 55.73   & 61.07   & 49.27   &&  71.74 &  70.77 \\
LSEP &annotation & \textbf{82.31}   & \textbf{83.07}  &  \textbf{81.76}  && \textbf{61.11}  & \textbf{65.33 }  &  \textbf{53.21}  && \textbf{72.80 }   & \textbf{73.97}  \\
\midrule 
MAttNet & image & 82.37   & 83.60 & 79.97  &&  68.05   &    70.41&  61.77  &&  74.75   & 73.44  \\
LSEP & image & \textbf{83.52} & \textbf{84.07}   & \textbf{81.90}  && \textbf{68.83}  & \textbf{71.77}    & \textbf{64.10}&&  \textbf{75.87}    & \textbf{75.92}   \\
\midrule 
MAttNet & category &\textbf{71.71}  & 69.68 &73.97 &&50.51  &44.00  &49.68  && 59.82 & 60.34\\
LSEP &category &71.64  & \textbf{72.90}   &\textbf{75.18} &&\textbf{51.73} &\textbf{47.33} & \textbf{52.91} &&  \textbf{63.90}  & \textbf{63.41} \\
\midrule
MAttNet & supercategory & \textbf{70.59} & 58.89  &\textbf{73.18} &&48.54  &\textbf{39.23}  &\textbf{50.27} &&  \textbf{56.88} & \textbf{54.31} \\
LSEP & supercategory &68.80  &\textbf{60.00}  &72.01&&  \textbf{48.61}  & 38.67 & 50.16&& 56.54 & 53.93 \\ 
\midrule
\end{tabular}
}
\resizebox{1.6\columnwidth}{!}
{
\begin{tabular}{p{1.45cm}p{2.8cm}<{\centering}p{0.6cm}<{\centering}p{0.6cm}<{\centering}p{0.6cm}<{\centering}p{0.0cm}<{\centering}p{0.6cm}<{\centering}p{0.6cm}<{\centering}p{0.6cm}<{\centering}p{0.0cm}<{\centering}p{0.6cm}<{\centering}p{0.6cm}<{\centering}}
\midrule
 F1&&\multicolumn{3}{c}{RefCOCO}
 && \multicolumn{3}{c}{RefCOCO+}
 && \multicolumn{2}{c}{RefCOCOg} \\ 
\cline{3-5}\cline{7-9}\cline{11-12}\\[-8pt]
score & type&Val & testA& testB &&Val& testA&testB &&Val& test  \\ \midrule
MAttNet & annotation & 35.30  & 34.59 & 44.40  &&  27.89  & 27.04   & 40.97   && 41.35  & 40.03\\
LSEP &annotation &  \textbf{37.03 } &\textbf{36.11}   &\textbf{49.25} &&\textbf{29.14 } & \textbf{28.27}  & \textbf{42.61}   &&  \textbf{43.10}    &\textbf{41.44}  \\
\midrule 
MAttNet & image &  38.21  & 37.50  & 43.16  &&   29.17  & 28.16   & 44.40   &&  43.23   & 42.75  \\
LSEP & image &   \textbf{39.87 } & \textbf{38.65}  & \textbf{56.51} &&  \textbf{30.02} &  \textbf{29.07}   &\textbf{45.83}     &&  \textbf{45.11 }   & \textbf{43.51}   \\
\midrule 
MAttNet & category &33.05  &  32.84 &34.89 & & 21.25 &15.38  & 22.17 & &25.91 &23.44\\ 
LSEP &category  & \textbf{36.41}  &\textbf{40.58}  &\textbf{35.89} &&\textbf{24.79} & \textbf{17.39} & \textbf{26.14} & &\textbf{32.86} &\textbf{32.08}  \\  
\midrule
MAttNet &supercategory &19.10  &  23.13&24.15&  &16.06 & 8.33 & 19.41 & &  10.08 & 12.77 \\
LSEP & supercategory&\textbf{22.83}&\textbf{31.25} &\textbf{25.32} & &\textbf{17.64} &\textbf{11.67} & \textbf{20.51}  &&\textbf{25.21}  & \textbf{26.37} \\ 
\bottomrule
\end{tabular}
}
\end{center}
\caption{Accuracy and F1 score with groundtruth bounding boxes provided by the MSCOCO dataset. `Annotation', `image', `category' and `supercategory' represent annotation-based, image-based, category-based and supercategory-based selections respectively. The ratio of fully-labeled query is set to be 50\% for all four settings.}
\label{table:zsl}
\end{table*}

\textbf{Category-based selection} is to select the phrases based on their categories. 
Half of the categories in the training set are annotated with their full queries, while only the category names along with the bounding boxes are available for the objects in the remaining categories. 

\textbf{Supercategory-based selection} is to select the queries based on their parent category. 
We either use the labeled queries for the objects in the same supercategory, or replace them with their category names during training.

For category-based and supercategory-based settings, attributes other than the category names are not available during training if only category names are used. 
We select 40 categories and 6 supercategories and label them with full queries during training, and the attribute loss will not be calculated for the entities without full queries.  
These 6 supercategories include person, accessory, sports, kitchen, furniture and electronic, and the selected 40 categories are those whose category IDs are not divisible by 2. 
The inference is conducted on the whole set for annotation and image-based settings, while only on the remaining 40 categories and 5 supercategories which are not selected for the category and supercategory-based selection. 
 
\subsection{Experimental Setup}\label{esss}
In this subsection, we start with the details for the pipeline using for training and inference, followed by the implementation details and evaluation metrics.

\textbf{Training and inference} During the training period, our model face two different types of data: labeled objects, whose bounding boxes are labeled with groundtruth phrases, and unlabeled objects, whose bounding boxes are labeled with incomplete or no annotations. 
We train our model for 50000 iterations in all.
We train the first 20000 iterations on labeled objects. 
After 20000 iterations, we initialize the two predictors for 5000 iterations with the grounding model. 
In the remaining iterations, we i) train the grounding model on labeled objects; ii) train our predictors with the assistance of the grounding model on labeled objects, and iii) use the predictors to generate the language embeddings and apply them to train the grounding model on unlabeled ones. 
We do these three steps recurrently. The learning rate is 1e-4 and decays to half after every 8000 iterations. 
For the unlabeled objects, we set $w_{subj} = w_{loc} = 0.5$ and $w_{rel} = 0$ when applying $\Tilde{q^{loc}}$ and $\Tilde{q^{subj}}$ as language embeddings.
During the inference, we evaluate the score for the grounding model following Eq.~\ref{mattnet1} and use the bounding boxes with the highest confidence as our final prediction for the given query. 
We follow MAttNet \cite{yu2018mattnet} use the bounding boxes from the groundtruth annotations of the MSCOCO dataset as candidate proposals for both training and inference for all methods.

\textbf{Implementation Details}
For visual feature extraction, we follow \cite{yu2018mattnet} and apply a ResNet-101-based \cite{he2016deep}  Faster RCNN \cite{girshick2015fast} to extract $v_{subj}$ and $v_{rel}$  for its appearing.
We first encode the original query word by word into one-hot vectors for language features, then extract the feature with a bi-LSTM.  
We extract the visual feature from the C3 and C4 layers from the same ResNet to build the subject embedding predictor. 
The subject feature is the concatenation of C3 and C4 features after two 1x1 convolutional kernels that do not share weights. 
The category feature is the C4 feature, followed by a 1x1 convolutional kernel. 
The visual embeddings for subjects and locations are sent into the subject and location embedding predictors respectively, 
which are both 2-layer MLPs with a 512-dim hidden layer. The activation function is set as ReLU for the pipeline.

\textbf{Metrics} 
We apply two metrics for evaluation: Accuracy for phrase grounding and F1 score for attribute prediction. 
The accuracy for phrase grounding is calculated as the percentage of correct bounding box predictions compared with the number of queries. 
The F1 score is the harmonic mean of the precision and recall for the attribute prediction. 
We follow \cite{kazemzadeh2014referitgame} to parse the queries into 7 parts: category name, color, size, absolute location, relative location, relative object, and generic attribute. 
We evaluate the F1 score on the three types of attributes handled by the subject embedding predictor: category name, color, and generic attributes. 
\subsection{Results and Analysis}\label{sec:res}
In this subsection, we first show quantitative results for four semi-supervised settings with different ratios of annotations, followed by the results using a detector instead of the groundtruth bounding boxes and some ablations about how much every module in LSEP contributes.

\textbf{Four-way characterization.}\label{semisup}
We show results for four different selection settings in Table~\ref{table:zsl}. 
For every case, 50\% of the objects have language queries, while the remaining 50\% only have category names. 
As shown in Table~\ref{table:zsl}, adding LSEP module to MAttNet improves accuracy for each of the annotation-based, image-based, and category-based selection settings. 
However, for the supercategory-based setting, it remains the same.
Comparing the results between category-based and supercategory-based selection settings, we show that the grounding model can apply features from a nearby category, even labels for such categories are not available, since the objects in the same supercategory share similar visual features. 

Interestingly, we find that F1 scores for all four settings significantly benefit from LSEP, 
indicating that the final attribute accuracy improves with our subject embeddings predictor. 
Better F1 results for all four selection settings show that LSEP enables better transfers of attribute combinations compared with MAttNet.

\textbf{Different ratios of available queries.}
We analyze the performance on different ratios of annotated queries in Table~\ref{table:ssl-ab}. 
We use 25\%, 50\%, 75\%, 100\% of the queries for annotation and image-based selection settings, where 100\%, i.e., every bounding box used for training is paired with a query, refers to the fully-supervised case. 
In this case, we first train the two predictors with groundtruth annotations. We then use these two language encoders to generate corresponding embeddings from the encoded visual embeddings and apply them for training the grounding model. 
We make the following observations. 
\begin{enumerate}
    \setlength{\itemsep}{0pt}
    \item[(i)] \textbf{Fully-supervised Results.} LSEP gives mild improvements over Mattnet when 100\% of the query annotations are available. 
    Compared with the fully supervised method, LSEP can extract further information besides information in the labeled queries. 
    
    \item[(ii)] \textbf{Different amount of annotations.} With fewer available queries in the training set, the performances for both MAttNet and LSEP go down, while LSEP shows consistent improvement compared with MAttNet on both annotation-based selection and image-based selection settings. LSEP can narrow the gap by around $40\%$ between the results of MAttNet to their supervised performance. 
    
    \item[(iii)] \textbf{Annotation density in the image.} Results for the image-based setting shows higher phrase grounding accuracy than the annotation-based selection setting with similar available queries, indicating that densely-annotated images help a grounding model find the differences between objects. 
    Objects in the same image create more challenging object-query pairs for the phrase grounding model to distinguish the subtle differences in the same surroundings and circumstances. 
    
    \item[(iv)] \textbf{On use of category names.} 
    Comparing the results of using the category names as the full queries during training, we notice that both MAttNet and LSEP show similar performance when not using them. 
    When using the category name as the full query, positive and negative examples can not be from the same category. 
    Distinguishing two objects in the same category is a more challenging task and helps the network find more useful descriptive information than finding the differences between two objects from different classes.
\end{enumerate}

\begin{table*}[]
\begin{center}
\resizebox{1.65\columnwidth}{!}
{
\begin{tabular}{p{2.5cm}p{2.65cm}<{\centering}p{0.6cm}<{\centering}p{0.6cm}<{\centering}p{0.6cm}<{\centering}p{0.0cm}<{\centering}p{0.6cm}<{\centering}p{0.6cm}<{\centering}p{0.6cm}<{\centering}p{0.0cm}<{\centering}p{0.6cm}<{\centering}p{0.6cm}<{\centering}}
\toprule
 &Type /&\multicolumn{3}{c}{RefCOCO}
 && \multicolumn{3}{c}{RefCOCO+}
 && \multicolumn{2}{c}{RefCOCOg} \\ 
\cline{3-5}\cline{7-9}\cline{11-12}\\[-8pt]
 & Labeled \% & Val  &  testA & testB&& Val &testA &testB && Val & testA   \\ \midrule  
 Accu-Att \cite{deng2018visual}&100\%& 81.27 & 81.17 & 80.01 && 65.56 & 68.76 & 60.63 &&  -&-\\ 
 PLAN \cite{zhuang2018parallel}&100\%&81.67 & 80.81 & 81.32 &&64.18 & 66.31& 61.46 &&  -&-\\ 
Multi-hop \cite{strub2018visual}&100\%& 84.90 & 87.40 & 83.10 &&73.80 & 78.70& 65.80 && -&-\\ 
NegBag \cite{nagaraja2016modeling} &100\%& 76.90&  75.60 & 78.00 && - & - & - &&  -&68.40 \\
S-L-R \cite{yu2017joint}  &100\%& 79.56 & 78.95 & 80.22 && 62.26 & 64.60 & 59.62 &&  71.65&71.92 \\
MAttNet \cite{yu2018mattnet}  & 100\%& 85.65 & 85.26 &  {84.57} && 71.01 & 75.13 & 66.17 &&  78.10 & 78.12\\ 
LSEP &100\%& {85.71}&  {85.69} & 84.26 && {71.99}&  {75.36}&  {66.25}&&  {78.96}& {78.29}\\ 
\midrule
MAttNet \textit{w}/\textit{o} cat.&annotation-75\%&81.78& 83.54 &79.26 &&61.08& 64.83& 56.58 && 72.14 & 72.07 \\ 
LSEP \textit{w}/\textit{o} cat. &annotation-75\%& {83.02}& {84.70} & {79.60} && {67.53}&  {70.07}&  {61.51 }&&  {75.53} &  {74.82} \\ 
MAttNet &annotation-75\%& 81.89  & 83.52     & 79.48    &&  61.72    &  64.87    & 56.53    && 72.30    & 72.02    \\  
LSEP &annotation-75\%&  83.11 &     84.46 &     79.58  &&  68.01    & 70.47     & 61.49    && 75.62    &    74.89 \\  
\midrule         
MAttNet \textit{w}/\textit{o} cat. &annotation-50\%&79.33& 80.07 &78.05 &&55.59& 60.95& 49.46 && 71.20& 70.75\\   
LSEP \textit{w}/\textit{o} cat. &annotation-50\%& {82.11}& {83.60} & {80.96} &&  {60.28}&  {65.19}&  {53.18} &&  {73.71}&  {72.93}\\ 
MAttNet &annotation-50\% &  79.18 & 80.05 & 78.71  && 55.73   & 61.07   & 49.27   &&  71.74 &  70.77 \\
LSEP &annotation-50\%& 82.31   & 83.07  &  81.76  && 61.07  & 65.33   &  53.21  &&  72.80    & 73.97  \\
\midrule         
MAttNet \textit{w}/\textit{o} cat. &annotation-25\%&63.73& 66.01 &62.88& &42.96& 48.10& 39.76 && 67.10 & 67.13 \\ 
LSEP \textit{w}/\textit{o} cat.  &annotation-25\%& {67.43}&  {70.55} & {66.01} & & {46.94} & {52.81} &  {41.42} &&  {69.08} &  {69.52} \\  
MAttNet &annotation-25\%&  63.02  &     65.59 &  62.93    &&    42.99  &    48.12  &    39.75 &&    67.21 & 67.09    \\  
LSEP  &annotation-25\%& 67.74   & 70.51     & 66.01    && 46.96     &    53.01  &  41.85  &&    69.06 & 69.58    \\        
\midrule
MAttNet \textit{w}/\textit{o} cat. &image-75\%& 82.89& 83.45 &81.51 &&68.40& 71.18& 64.14 & & 75.76  & 75.32  \\      
LSEP \textit{w}/\textit{o} cat. &image-75\%& {83.70}&  {84.38} & {82.85} && {70.12}&  {73.05}&  {64.59} &&  {75.94} &  {76.15} \\  
MAttNet &image-75\%&  82.87 &  84.01    & 81.39    &&  68.47    &  71.09    & 64.18    && 75.99   & 75.26    \\  
LSEP &image-75\%&  83.91  &     84.16 & 82.87    &&     70.09 & 72.98     & 64.61    && 76.27    & 76.09    \\  
\midrule
MAttNet \textit{w}/\textit{o} cat.&image-50\%&82.40& 83.33 &80.79 &&67.42& 70.19& 61.96 && 74.61 & 73.65\\  
LSEP \textit{w}/\textit{o} cat.&image-50\%& {83.34}&  {83.54} & {82.16} && {68.85}&  {71.08}&  {63.31} &&  {75.94}&  {75.35}\\ 
MAttNet&image-50\%& 82.37   & 83.60 & 79.97  &&  68.05   &    70.41&  61.77  &&  74.75   & 73.44  \\
LSEP &image-50\% & 83.52 & 84.07   & 81.90  && 68.83  & 71.77    & 64.10&&  75.87    & 75.92   \\
\midrule             
MAttNet \textit{w}/\textit{o} cat.&image-25\%& 79.92 & 80.25   &78.81 && 63.90 & 66.92   &59.85 &&68.34  &68.20\\ 
LSEP \textit{w}/\textit{o} cat. &image-25\%&  {82.32} &  {82.77}&  {80.48 } & & {67.22} &  {69.80}  &  {62.38}& &  {73.14} & {72.42}\\  
MAttNet &image-25\%& 79.81 & 80.59  &    78.76 &&  63.95  & 66.92    & 59.81    &&    68.40     &  68.07    \\ 
LSEP&image-25\%& 82.75    & 82.53   & 81.02   &&    67.19 & 69.97    & 62.53    &&   73.25  &    72.08 \\  
\bottomrule
\end{tabular}}
\end{center}
\caption{Semi-supervised results for grounding for annotation-based and image-based selections. `Annotation' and `image' represent annotation-based and image-based selections, and the number after dash represent the percentage of bounding boxes that has been labeled with queries during training. {Methods ending with ``\textit{w}/\textit{o} cat." do not use category names as labeled queries.}}
\label{table:ssl-ab}
\end{table*}

\textbf{Results of Using Detection Outputs} 
Instead of using MSCOCO groundtruth boxes, we compare the semi-supervised phrase grounding results with the proposals generated by a pretrained detector. We use the image-based selection as the experiment setting, and set the percentage of images that have been annotated as 50\% for all configurations. We use a Faster R-CNN \cite{ren2015faster} trained on MSCOCO 2014 \cite{lin2014microsoft} detection task provided by MAttNet \cite{yu2018mattnet} as our detector. During training, we use the proposal with the maximum IoU with the groundtruth bounding box for those objects with labeled queries. For the remaining 50\% that are unlabeled with any query, we use the detection results and their category names provided by the detector. For inference, we select among the detected bounding boxes generated by the Faster R-CNN. 

We show the grounding accuracy on all three datasets in Table~\ref{table:detect} with the fully supervised results for comparison, where all queries are available. 
By comparing the gap between supervised results of MAttNet and the settings when the number of labeled bounding boxes is limited, we find that the grounding accuracy is relatively 34.9\% better on average and LSEP shows consistent improvement for all three datasets when applying it for predicting the query embeddings, indicating that we can apply LSEP for extracting information to learn from the images without any annotations in addition to the labeled proposals. 

\begin{figure*}
    \centering\resizebox{.965\linewidth}{!}{
    \begin{tabular}{cccc}
    
        \includegraphics[width=0.25\linewidth]{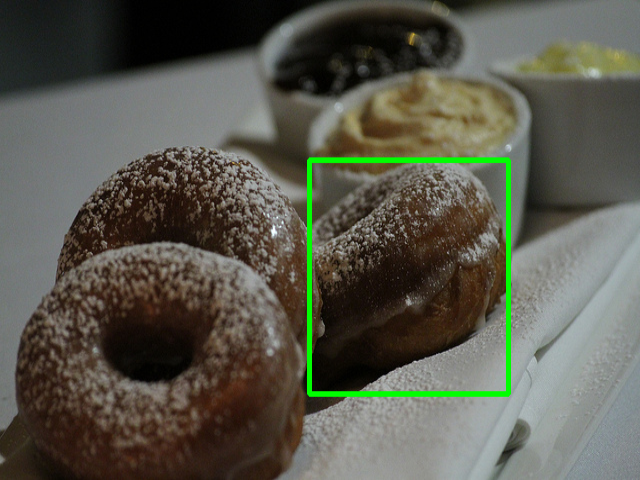} & 
        \includegraphics[width=0.25\linewidth]{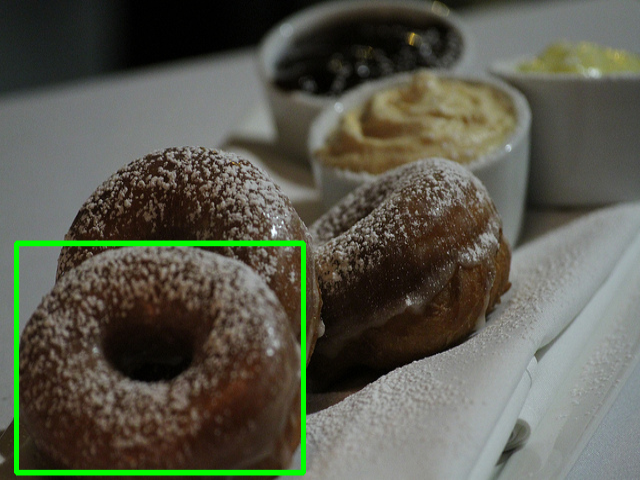} & 
        \includegraphics[width=0.25\linewidth]{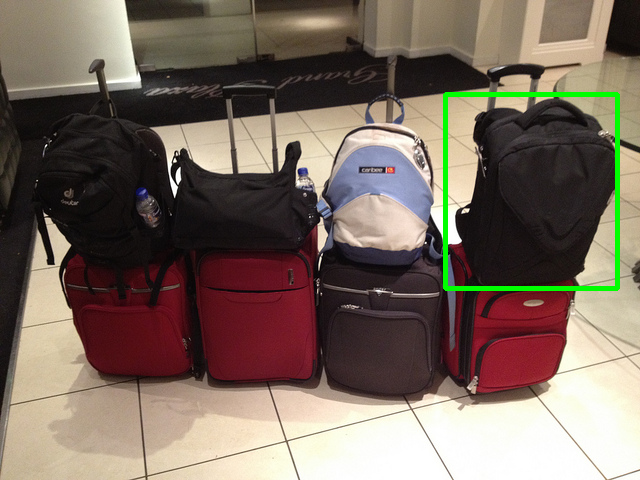} & 
        \includegraphics[width=0.25\linewidth]{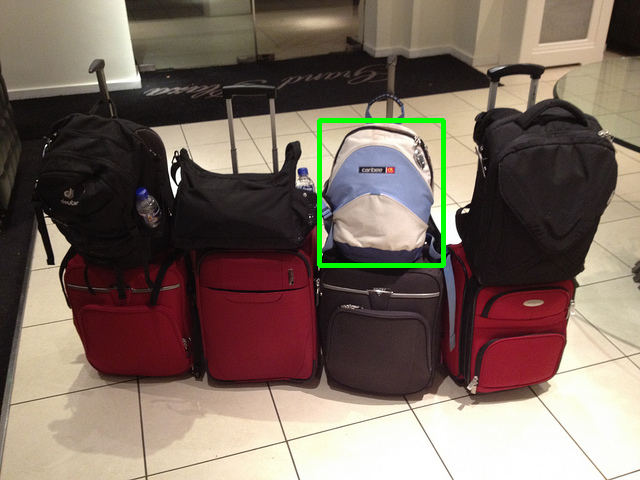} 
        \\
        \multicolumn{2}{c}{(a) The donut in the front} & \multicolumn{2}{c}{(b) A white and blue bag on top of a black suitcase}\\
        
       \includegraphics[width=0.25\linewidth]{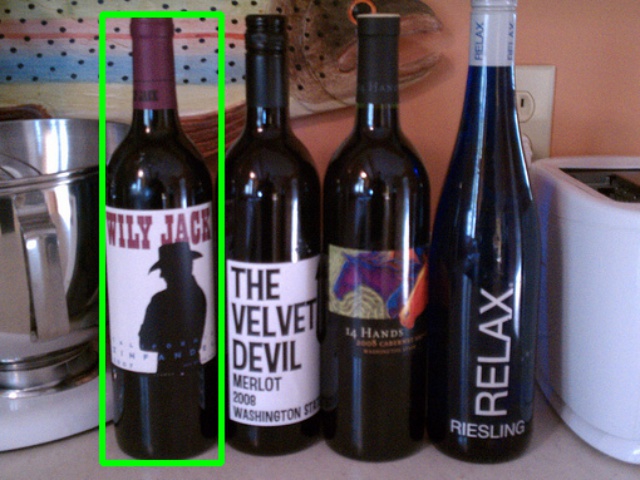} & 
        \includegraphics[width=0.25\linewidth]{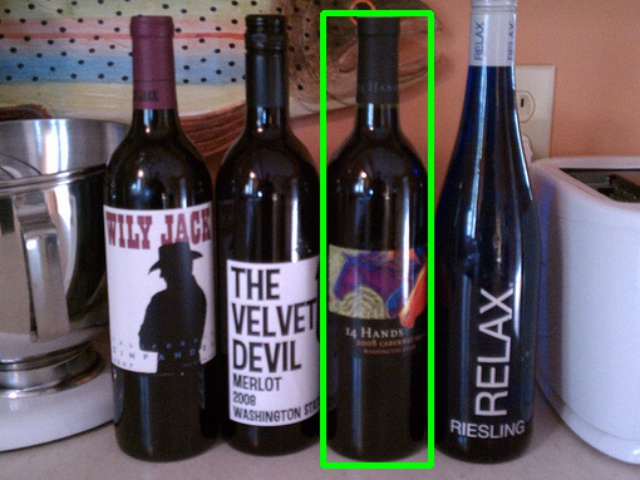} & 
        \includegraphics[width=0.25\linewidth]{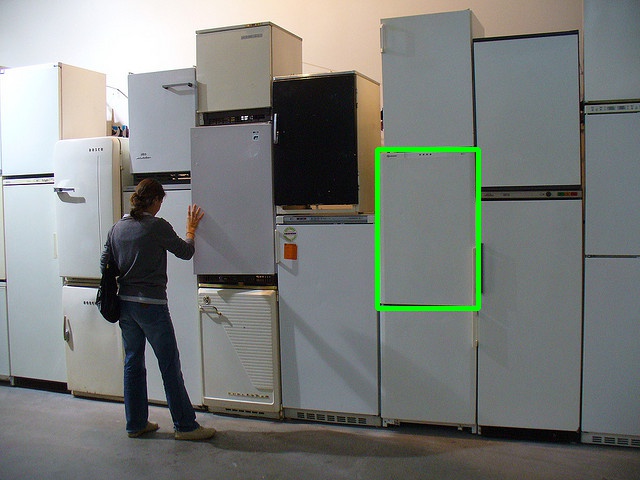} & 
        \includegraphics[width=0.25\linewidth]{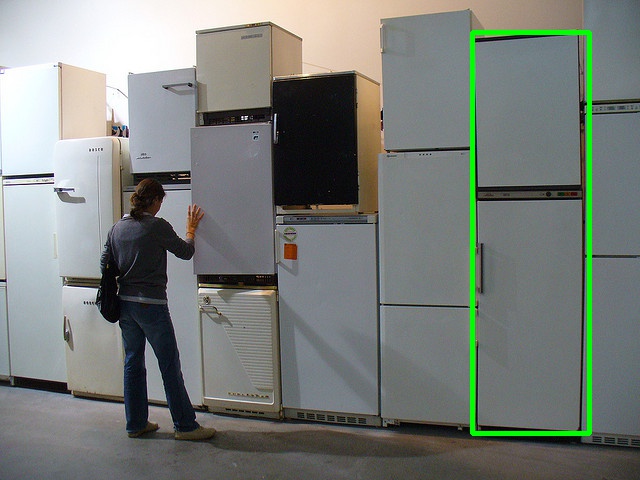} 
        \\
        \multicolumn{2}{c}{(a) Third bottle from the left} & \multicolumn{2}{c}{(b) Second stack of $<$UNK$>$ from right}
        \\
        
        \includegraphics[width=0.25\linewidth]{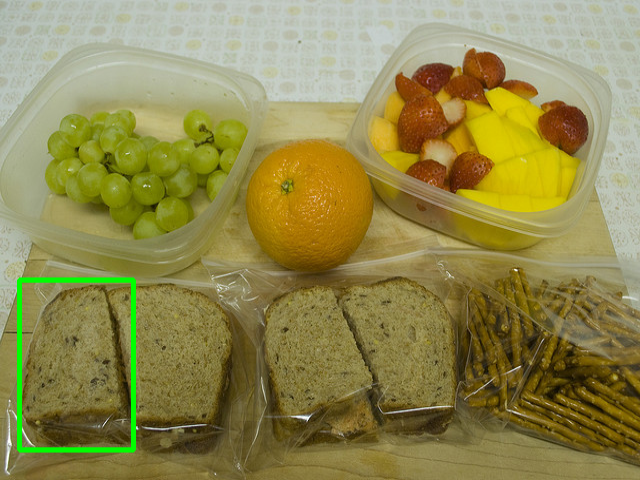} & 
        \includegraphics[width=0.25\linewidth]{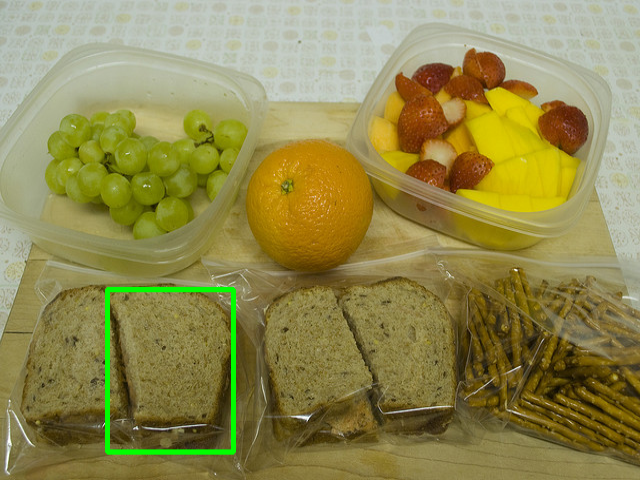} & 
        \includegraphics[width=0.25\linewidth]{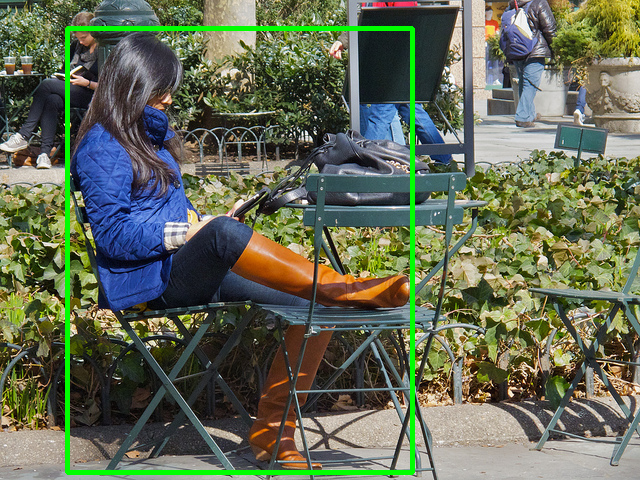} & 
        \includegraphics[width=0.25\linewidth]{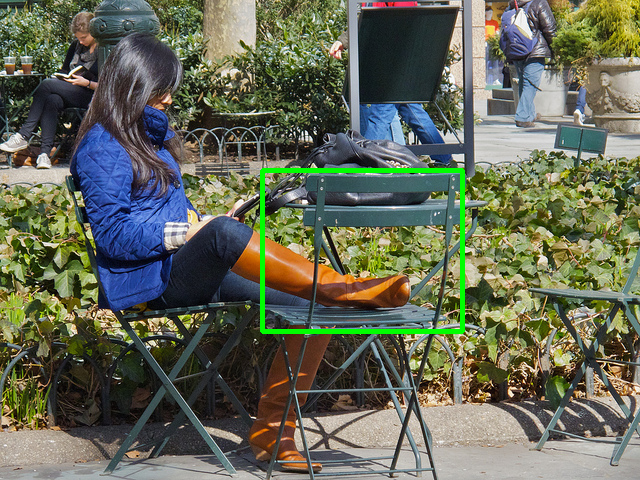} 
        \\
        \multicolumn{2}{c}{(c) The right half of the left sandwich} & \multicolumn{2}{c}{(d) Chair her foot is on}
        \\

        \includegraphics[width=0.25\linewidth]{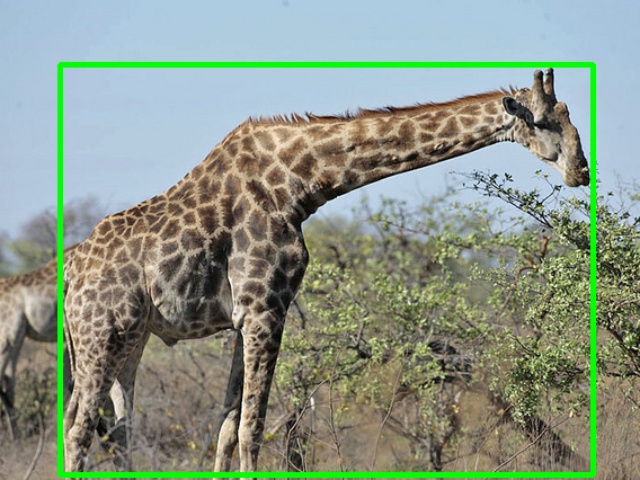} & 
        \includegraphics[width=0.25\linewidth]{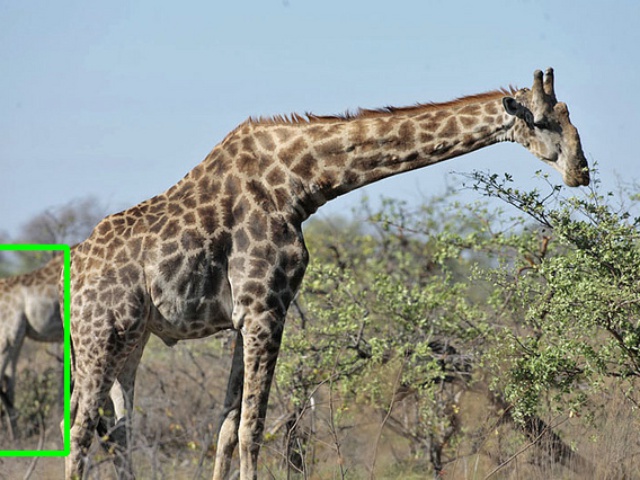} & 
        \includegraphics[width=0.25\linewidth]{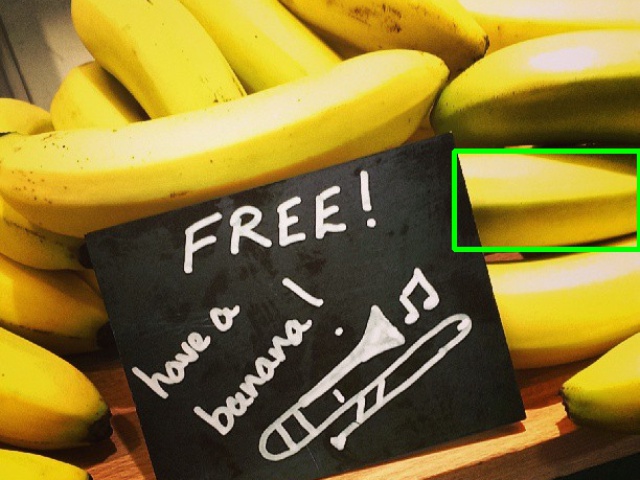} & 
        \includegraphics[width=0.25\linewidth]{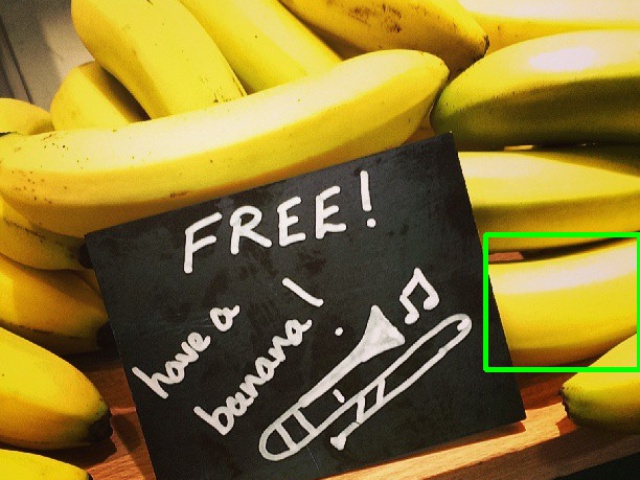} 
        \\
        \multicolumn{2}{c}{(c) The giraffe whose head is not visible} & \multicolumn{2}{c}{(d) Right side second banana up}
        \\

    \end{tabular}%
     }
    
    \par\bigskip
\caption{Visualization results. The images on the left are the results for MAttNet and images on the right are the results for LSEP.}
\label{fig:vis}
\end{figure*}

\textbf{Contribution of Location and Subject Modules} 
We conduct an ablation study on how much the two embedding predictors contribute to the improvement of semi-supervised grounding accuracy.
We use the RefCOCO dataset with 50\% annotations available based on annotation-based selection as our setting. 
We show the results in Table~\ref{table:ablation}. We observe that both the subject and location embedding predictors improve the grounding accuracy compared with the model without any embedding predictor. The combination of two predictors has the highest score.
Both subject and object predictors help with better grounding results compared with original MAttNet.

\subsection{Qualitative Results}\label{qualitative}
We show some visualization results in Fig.~\ref{fig:vis}. 
Results in the four rows are for annotation-based, image-based, category-based, and supercategory-based selection settings respectively. 
We show the grounding results for MAttNet of the same pair on the left and LSEP on the right.
We calculate the results for the category-based and supercategory-based selection settings on the categories whose full queries are not available during training.
We find that MAttNet successfully finds the object of the same category, such as the `bottle' in the first example of the second row, but it fails to find the one based on the given query, while LSEP successfully localizes the third bottle from the left in the image.

\begin{table}[]
\begin{center}
\resizebox{.95\columnwidth}{!}
{
\begin{tabular}{p{1.7cm}p{.9cm}p{2.2cm}<{\centering}p{1.3cm}<{\centering}p{1.3cm}<{\centering}}
\toprule
Dataset & Split & MAttNet (FS) & MAttNet   & LSEP \\
 \midrule
RefCOCO & val   & 75.78 & 73.17 & 74.25 \\
        & testA & 82.01 & 79.54 & 80.47 \\
        & testB & 70.03 & 67.83 & 68.59\\  
 \midrule
RefCOCO+ & val   & 65.88 & 63.95 & 64.67 \\
         & testA & 72.02 & 69.33 & 70.25 \\
         & testB & 57.03 & 53.60 & 55.01 \\
\midrule
RefCOCOg & val   & 66.87 & 63.29 & 64.28 \\
         & test  & 67.03 & 62.97 & 64.01\\  
\bottomrule
\end{tabular}}
\end{center}

\caption{Phrase grounding ccuracy with Fast RCNN detector on image-based selection with 50\% fully annotated queries. MAttNet (FS) refers to the fully supervised result where 100\% labeled queries are available.}
\label{table:detect}
\end{table}

\section{Conclusion}\label{sec:con}
We study the task of using unlabeled object data to improve the accuracy of a phrase grounding system by using embedding predictor modules. This approach also allows us to introduce new categories, with available detectors, in phrase grounding. We show the improvements in accuracy by using subject and location embedding predictors applied to MAttNet \cite{yu2018mattnet} for semi-supervised grounding tasks.

\begin{table}[]
\begin{center}
\resizebox{.95\columnwidth}{!}
{
\begin{tabular}{p{1.cm}p{1.3cm}<{\centering}p{1.3cm}<{\centering}p{1.3cm}<{\centering}p{1.5cm}<{\centering}}
\toprule
Method &  MAttNet & SEP-Net & LEP-Net  & LSEP \\
 \midrule
val&  79.18 
& 80.85& 80.98&{82.31}\\      
testA&80.05
&82.09&81.81 &{83.07}\\          
testB  & 78.71 
&79.57&78.88&{81.76}\\  
\bottomrule
\end{tabular}}
\end{center}

\caption{Ablation study. LEP-Net only uses the location embedding predictor and SEP-Net uses the subject embedding predictor.}
\label{table:ablation}
\end{table}

\section*{Acknowledgments}
This work was supported by the U.S. DARPA AIDA Program No. FA8750-18-2-0014. The views and conclusions contained in this document are those of the authors and should not be interpreted as representing the official policies, either expressed or implied, of the U.S. Government. The U.S. Government is authorized to reproduce and distribute reprints for Government purposes notwithstanding any copyright notation here on.

{\small
\bibliographystyle{ieee_fullname}
\bibliography{egbib}
}

\end{document}